\documentclass{article}
\usepackage{textcomp}
\usepackage{spconf,amsmath,graphicx,hyperref}
\usepackage[utf8]{inputenc}
\usepackage{makecell}

\usepackage{enumitem}
\usepackage{microtype}
\usepackage{balance}
\usepackage{graphicx}
\usepackage{booktabs} % 可选：用于更漂亮的表格线（如\toprule, \midrule）
\usepackage{multirow} % 用于合并行
\title{PankRAG: Enhancing Graph Retrieval via Globally Aware Query Resolution and Dependency-Aware Reranking Mechanism}
\name{Ningyuan Li$^{1}$,
      Junrui Liu$^{1}$,
     Yi Shan$^{1}$,
     Minghui Huang$^{1}$,
     Ziren Gong$^{2}$,
     Tong Li$^{1}$%
\sthanks{Corresponding author}}
\address{
$^{1}$ College of Computer Science, Beijing university of technology, Beijing, China\\
$^{2}$ Department of Computer Science and Engineering, Bologna University, Bologna, Italy}
\begin{document}

\ninept
\maketitle
\begin{abstract}
% The abstract should appear at the top of the left-hand column of text, about
% 0.5 inch (12 mm) below the title area and no more than 3.125 inches (80 mm) in
% length.  Leave a 0.5 inch (12 mm) space between the end of the abstract and the
% beginning of the main text.  The abstract should contain about 100 to 150
% words, and should be identical to the abstract text submitted electronically
% along with the paper cover sheet.  All manuscripts must be in English, printed
% in black ink.
% Recent graph-based RAG approaches have improved retrieval performance by first extracting entities from queries and then retrieving their associated relationships and metadata from pre-constructed knowledge graphs. 
Recent graph-based RAG approaches leverage knowledge graphs by extracting entities from a query to fetch their associated relationships and metadata.
However, relying solely on entity extraction often results in the misinterpretation or omission of latent critical information and relationships. This can lead to the retrieval of irrelevant or contradictory content, as well as the exclusion of essential information, thereby increasing hallucination risks and undermining the quality of generated responses. In this paper, we propose PankRAG, a framework designed to capture and resolve the latent relationships within complex queries that prior methods overlook. It achieves this through a synergistic combination of a globally-aware hierarchical resolution pathway and a dependency-aware reranking mechanism. PankRAG first generates a globally aware resolution pathway that captures parallel and progress relationships, guiding LLMs to resolve queries through a hierarchical reasoning path. Additionally, its dependency-aware reranking mechanism utilizes resolved sub-question dependencies to augment and validate the retrieved content of the current unresolved sub-question. Experimental results demonstrate that PankRAG consistently outperforms existing state-of-the-art methods, underscoring its generalizability. The code is available at:\url{https://github.com/Ninggggy/PankRAG}.

% In this paper, we propose PankRAG, a framework designed to capture and resolve the latent relationships within complex queries that prior methods overlook. It achieves this through a synergistic combination of a globally-aware hierarchical resolution pathway and a dependency-aware reranking mechanism.

% In this paper, we propose PankRAG, which resolves queries through a hierarchical resolution pathway with global awareness, augmented by a novel dependency-aware reranking mechanism.
\end{abstract}


\section{Introduction}
Large language models (LLMs) ~\cite{minaee2024large,naveed2025comprehensive}exhibit exceptional performance across diverse tasks by leveraging pre-training \cite{hoffmann2022training} on massive text corpora to generate appropriate responses. While LLMs' capabilities can be further enhanced through fine-tuning \cite{ding2023parameter} on downstream tasks, finite parameter capacity limits their ability to store comprehensive and evolving world knowledge. Retrieval-augmented generation (RAG) ~\cite{Lewis2020ragnlp,Guu2020REALM,gao2023precise,ram-etal-2023-context,Fan2024surveyrag,Izacard2023Atlas,gao2024retrievalaugmentedgenerationlargelanguage,kang2026multimodalmultiagentempoweredlegal}has emerged as the predominant solution, owing to its straightforward implementation. It enables static models to access up-to-date external knowledge while preserving interpretability and provenance tracking.

RAG dynamically integrate heterogeneous knowledge bases~\cite{sun2023think, edge2024local,guo2024lightrag,ma2024think,jimenez2024hipporag} to enhance contextual reasoning and factual grounding. Among these, knowledge graphs ~\cite{chen2020kgreview} serve as structured frameworks that encode information through interconnected entities and relationships, explicitly modeling semantic dependencies to enable human-like understanding and multi-hop logical inference. 
% Retrieval-Augmented Generation (RAG) enhances model reasoning and grounding by integrating external knowledge~\cite{sun2023think, edge2024local,guo2024lightrag,ma2024think,jimenez2024hipporag}. Knowledge graphs (KGs) are particularly effective for this, as their structure of interconnected entities and relationships explicitly models semantic dependencies, thereby enabling multi-hop logical inference.Recently, many approaches (including HippoRAG\cite{jimenez2024hipporag} and G-Retriever\cite{he2024g}) have integrated RAG with knowledge graphs. 
Recently, many approaches have integrated RAG with knowledge graphs to enhance factual grounding and multi-hop inference, with prominent examples including GraphRAG\cite{edge2024local}, LightRAG\cite{guo2024lightrag}, HippoRAG\cite{jimenez2024hipporag}, and G-Retriever\cite{he2024g}. For instance, GraphRAG\cite{edge2024local} enhances question answering by adaptively synthesizing precomputed summaries of entity communities. LightRAG\cite{guo2024lightrag} leverages a hierarchical retrieval system that fuses graph-based reasoning with vector matching to ensure efficient knowledge extraction and semantic fidelity for multi-hop queries. However, the single-phase retrieval approach of current graph-based RAG methods is ill-suited for deep reasoning, causing them to misinterpret or omit latent critical information and relationships when handling complex queries. For Specific Complex Queries (SCQ) \cite{guo2024lightrag} requiring precise multi-hop inference, they often retrieve broad, irrelevant information centered on a core entity instead of the specific relational facts, increasing the risk of LLMs' hallucinations. Similarly, for Abstract Complex Queries (ACQ) \cite{guo2024lightrag} demanding conceptual synthesis, they tend to fetch superficial definitions while overlooking the rich, implicit context required for a nuanced response.

In this paper, we propose PankRAG, which resolves complex queries through a hierarchical reasoning pathway with global awareness, augmented by a dependency-aware reranking mechanism.
Specifically, our approach leverages the robust planning capabilities of LLMs to generate a globally-aware (i.e., holistically planned rather than stepwise) resolution pathway, explicitly revealing parallel and progress relationships among sub-questions. Next, we refine complex queries by integrating the implicit critical information resolved from prior sub-question resolutions. This dual-level process produces a hierarchical resolution pathway with global awareness, significantly enhancing retrieval
performance and response quality.
It enables LLMs to dynamically disambiguate and contextualize ambiguous or polysemous queries before final answer generation.
% During execution, when processing a sub‑question that depends on prior sub-questions, our dependency‑aware reranking mechanism augments and reorders retrieved content by integrating these resolved sub-question dependencies. 
During execution, our dependency-aware reranking mechanism augments and validates retrieved content for sub-questions.
This process refines the retrieved context by integrating resolved sub-question dependencies, thereby enhancing the alignment between retrieved contexts and resolved sub-question dependencies.

% We evaluate PankRAG on four benchmark datasets for ACQ and two for SCQ, demonstrating that it consistently outperforms state‑of‑the‑art baselines across six benchmarks. 

% Compared to NaiveRAG, GraphRAG, and LightRAG, PankRAG achieves average win rates of 87.0\%, 59.78\% and 61.59\% respectively in ACQ datasets, and achieves average metric improvements of 17.42\%, 11.61\%, 11.08\% in SCQ datasets.

 Our contributions to this work are as follows:
\begin{itemize}[itemsep=2pt,topsep=0pt,parsep=1pt]
     \item 
     % We propose a hierarchical query decomposition strategy that systematically plans globally-aware problem-solving pathways, enhancing LLMs' capacity to capture and resolve implicit information and relationships, fostering alignment between query intent and generated responses.
     % We propose a hierarchical query decomposition strategy that generates a globally-aware resolution pathway, enhancing LLMs' capacity to resolve latent relationships, better align generated responses with query intent.
     We propose a hierarchical query decomposition strategy that generates a globally-aware resolution pathway, enabling LLMs to capture and resolve latent information and relationships, ensuring a tight alignment between generated responses and query intent.
     \item We propose a dependency-aware reranking mechanism that augments and validates retrieved content for the current sub-question by integrating resolved sub-question dependencies.
     \item  PankRAG demonstrates consistent superiority over state-of-the-art baselines, achieving comprehensive improvements on four ACQ datasets evaluated across six dimensions and two SCQ datasets assessed via four metrics.
 \end{itemize}

     % During Globally-Aware Planning, queries are decomposed into colored resolution pathways. In the bottom-up phase, two parallel sub-question sequences are generated: a \textcolor{blue}{blue} sequence for interdependent sub-questions that require sequential processing, and an \textcolor{green}{green} independent sequence that can be resolved concurrently. In the top-down phase, an \textcolor{orange}{orange} sequence rewrites complex queries, incorporating resolved sub-question dependencies to enhance clarity.
\begin{figure*}
    \centering
    % \vspace{-0.6cm}
    % \setlength{\abovecaptionskip}{2cm}
\includegraphics[width=0.88\linewidth]{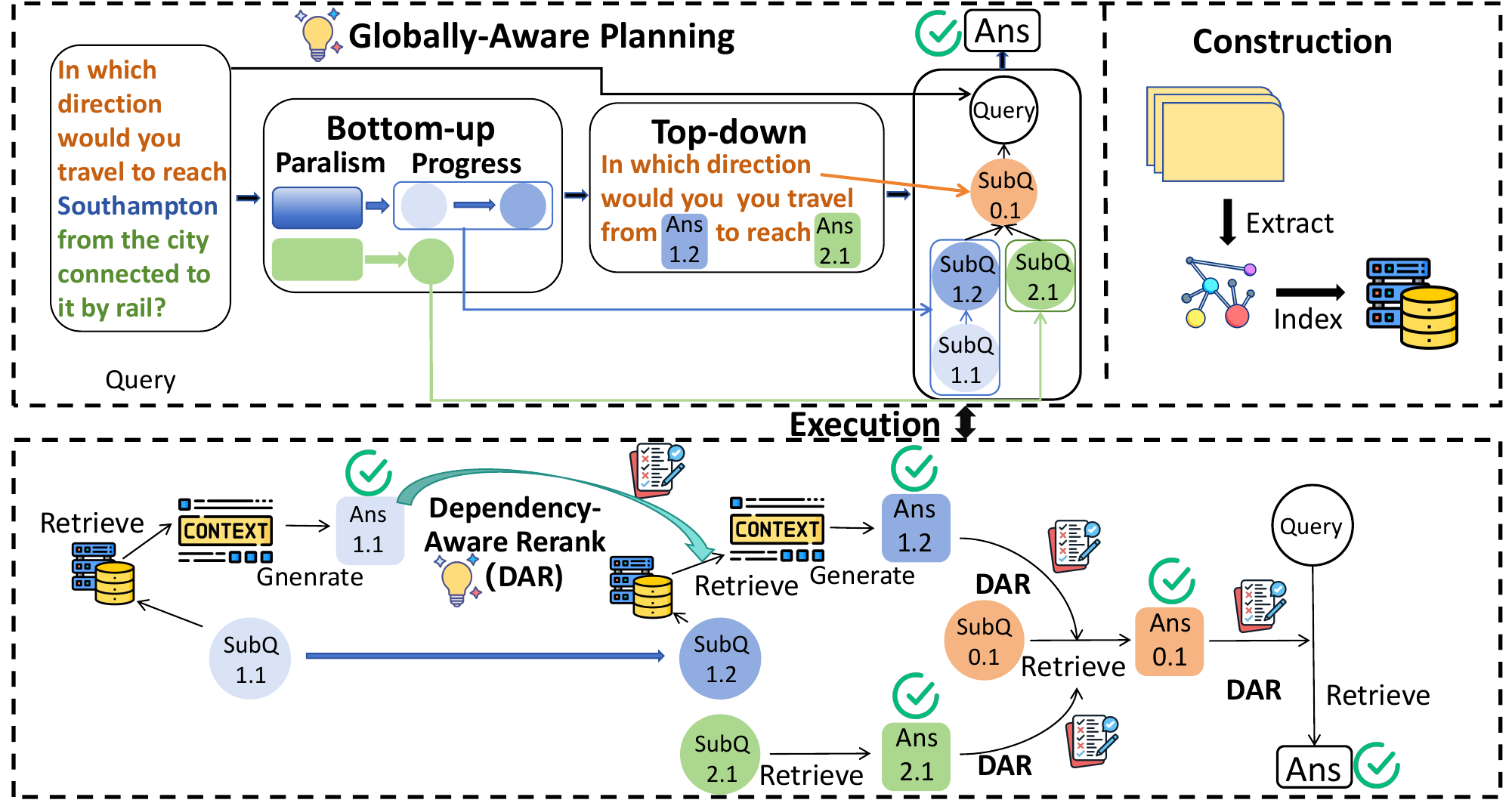}
    \vspace{-0.4cm}
    \caption{The overall workflow of PankRAG.}
    \label{fig:approach}
    \vspace{-0.6cm}
\end{figure*}

\section{Methodology}
% In this section, we detail our proposed framework, which comprises three core components. First, \textbf{Graph Construction} extracts entities and relationships from external knowledge sources to build a structured knowledge graph. Next, \textbf{Globally-Aware Planning} formulates a query resolution pathway by identifying latent critical information and relationships, improving LLMs' comprehensiveness of complex queries. Finally, during \textbf{Execution}, the LLM processes sub-questions either concurrently or sequentially based on their interrelationships. After retrieving context for a sub-question, the \textbf{dependency-aware reranking mechanism} augments and validates the content when the current sub-question depends on a prior sub-question. The LLM then leverages this refined context to generate accurate responses. The overall workflow of FG-RAG is shown in \autoref{fig:approach}.

 \subsection{Graph Contruction}

The source document corpus $D$  is segmented into $n$ text chunks 
$ \{D_i\}^n_{i=1} $. As the \autoref{equ:chunk}, for each chunk $ D_i $, the LLM performs entity recognition and relationship extraction, abstracting entities $ V_i $ and relations $ E_i $ from unstructured text via prompting strategies. A knowledge graph $ G $ is then constructed by instantiating nodes representing entities $ \{V_i\} $ and edges representing relationships $ \{E_i\} $ in \autoref{equ:graph}.
\begin{equation}
% Document chunking
% \setlength{\abovedisplayskip}{3pt}
(\mathcal{V}_i, \mathcal{E}_i) = \bigcup_{D_i \in \mathcal{D}} \textit{Extract}(D_i)
\label{equ:chunk}
\end{equation}
\begin{equation}
% Document chunking
\mathcal{G} = \left\{ \mathcal{V}_i, \mathcal{E}_i\right\}_{i=1}^n
\label{equ:graph}
\end{equation}
Next, Multiple communities $ \{Com_i\} $ are aggregated into a higher-level community  $ H_i $ via Leiden clustering, and this process is iteratively applied to hierarchically construct a multi-layered community structure. Finally, multi‑scale community summarization is executed by the LLM.
% , which populates a summary template with various elements (nodes, edges, and associated claims).
% \begin{equation}
% % Document chunking
% (\mathcal{V}_i, \mathcal{E}_i) = \bigcup_{D_i \in \mathcal{D}} \textsc{Extract}(D_i)
% \end{equation}
\begin{equation}
% Community detection
\mathcal{H}_i = \textit{Leiden}((\textit{Com}_i)_{i=1}^k)
\label{equ:leiden}
\end{equation}
% \begin{equation}
% % Entity-relationship extraction
% \mathcal{S}_i = \textit{Summarize}(\mathcal{H}_i)
% \end{equation}
% \subsubsection{Dependency-Aware Query Resolution Pathway Generation}
\subsection{Globally-Aware Planning}
\label{sec:plan}
\begin{figure}[ht!]
    \vspace{-0.6cm}
    \includegraphics[width=0.96\linewidth]{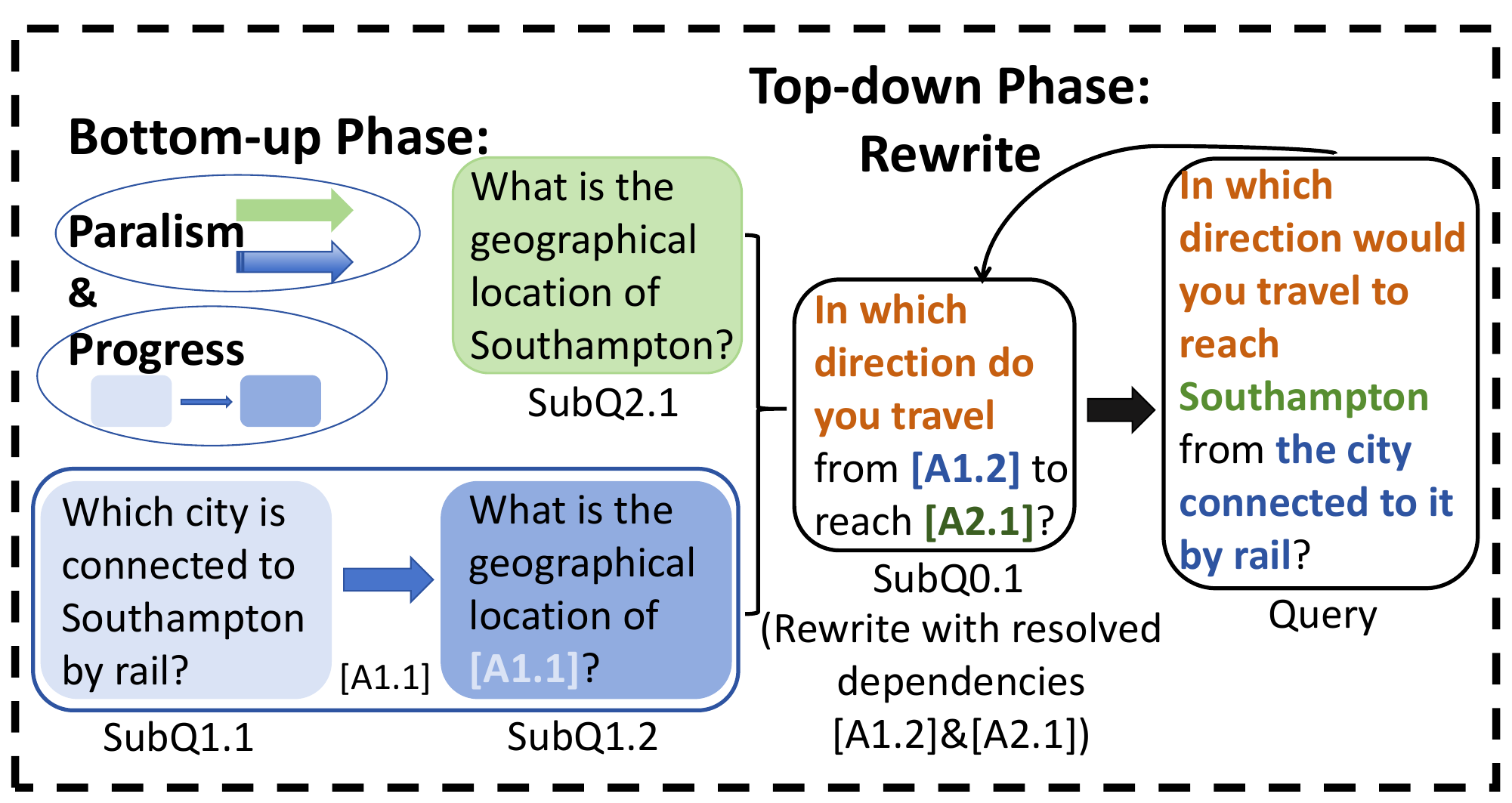}
    \vspace{-0.4cm}
    \caption{Operation example of Globally-Aware Planning. [A1.1], [A1.2], and [A2.1] correspond to the resolved outputs of SubQ[1.1], SubQ[1.2], and SubQ[2.1], respectively.}
    \label{fig:plan_example}
    % \vspace{-0.6cm}
\end{figure}
We propose a hierarchical query decomposition strategy that synergizes bottom-up relationship decomposition with top-down contextual rephrasing to address both implicit parallel or progress relationships and ambiguous or polysemous expressions in complex queries. 
Unlike conventional iterative approaches (e.g., Chain-of-Thought, CoT), which suffer from fixed iteration counts and myopic decomposition prone to local optima, our strategy dynamically adjusts the granularity and sequence of reasoning steps based on the query's inherent structure. 

% In the bottom‑up phase, two crucial relationship types are detected. 
% First, parallelizable task relationships are identified to enable concurrent resolution of mutually independent sub‑questions, optimizing computational efficiency by reducing redundant resource contention and minimizing latency. Second, latent dependencies are systematically extracted to ensure prerequisite sub-questions are solved before dependent tasks. 
% Simultaneously, the top‑down phase disambiguates polysemous expressions into logically coherent substructures, combining the resolved sub-question dependencies. This dual mechanism reduces the risk of deviation in reasoning paths and enhances computational efficiency by eliminating redundant iterations, achieving a balanced trade‑off between precision and scalability. The detailed workflow is presented below, and a real-world query in Multihop-RAG dataset example is above in \autoref{fig:plan_example}.

\begin{itemize}[itemsep=0pt,topsep=0pt,parsep=0pt]
    % \item Parallel Relationship Detection
    % This step is used to identify parallel semantic relationships, such as independent sub-questions or co-occurring entities. If no parallelism is detected, a single resolution sequence (ID=1) is initialized. When parallelism exists, multiple concurrent sequences (IDs $\geq$1) are adaptively spawned, leveraging dependency tree analysis and semantic role labeling to optimize concurrency efficiency.

    % \item Implicit Dependency Resolution 
    % Latent dependencies (e.g., causal prerequisites, temporal constraints) are extracted from the query’s contextual and lexical cues. These dependencies are formalized into a directed acyclic graph (DAG), where nodes represent sub-questions and edges encode dependency relationships. Topological sorting is applied to enforce prerequisite-resolution order, ensuring logical coherence during execution.
    \item \textbf{Bottom-up phase}:
    % The Parallel Relationship Detection step is used to identify parallel semantic relationships, such as independent sub-questions or co-occurring entities. If no parallelism is detected, a single resolution sequence (ID=1) is initialized. When parallelism exists, multiple concurrent sequences (IDs $\geq$1) are adaptively spawned, leveraging dependency tree analysis and semantic role labeling to optimize concurrency efficiency. Furthermore, Latent dependencies (e.g., causal prerequisites, temporal constraints) are extracted from the query’s contextual and lexical cues. These dependencies are formalized into a directed acyclic graph (DAG), where nodes represent sub-questions and edges encode dependency relationships. Topological sorting is applied to enforce prerequisite-resolution order, ensuring logical coherence during execution.
    In the Parallel Relationship Detection step, parallel semantic relationships—such as independent sub‑questions or co‑occurring entities—are identified  to enable parallel execution, thereby optimizing computational efficiency by minimizing resource contention. If no parallelism is detected, a single resolution sequence is initialized. 
    When parallelism exists, multiple concurrent sequences (SubQ1.x and SubQ2.x in \autoref{fig:plan_example}) are adaptively spawned to optimize concurrent execution. Concurrently, latent progress relationships are extracted from the query. Upon identifying distinct relationships within the query, sub-questions are generated and formalized into a directed acyclic graph (DAG), where nodes represent sub‑questions and edges encode their relationships. Finally, topological sorting is applied to enforce prerequisite‑resolution order, ensuring logical coherence during execution.
    \item \textbf{Top-down phase}:
    For queries containing explicit ambiguities or polysemous expressions, a disambiguation sequence combining resolved sub-question dependencies (SubQ0.x in \autoref{fig:plan_example}) is invoked. In this phase, an LLM-driven rephrasing process converts ambiguous queries into logically partitioned formulations. 
\end{itemize}

This dual mechanism reduces the risk of deviation in reasoning paths and enhances computational efficiency by eliminating redundant iterations, achieving a balanced trade‑off between precision and scalability. The detailed workflow is above in\autoref{fig:approach}, and a real-world query in the Multihop-RAG \cite{tang2024multihop} dataset example is presented in \autoref{fig:plan_example}.
\subsection{Execution}
% For the retrieval step, we adopt GraphRAG\cite{edge2024local}'s dual approach: employing local search for SCQs and global search for ACQs.
% \subsubsection{Retrieval}
Following GraphRAG\cite{edge2024local}, we tailor our retrieval strategy to the query type. For SCQs, we employ a local search that targets query-referenced nodes and expands to their neighbors, ensuring precise entity-centric grounding. For ACQs, we utilize hierarchical community retrieval, dynamically selecting between detailed leaf-level and thematic high-level summaries to balance retrieval breadth and depth.
\begin{table}[!t]
\centering
\caption{Overall Performance on the ACQ Datasets.}
\label{tab:abstract_res}
\resizebox{1\columnwidth}{!}{  % ← 缩小到 80% 页宽
{ % 开始一个局部作用域
\renewcommand{\arraystretch}{0.3} % 将行高缩放为原来的85%
\setlength{\tabcolsep}{0.5pt}
% \normalsize
\begin{tabular}{lcccccccc}
\toprule
& \multicolumn{2}{c}{\textbf{Agriculture}}
& \multicolumn{2}{c}{\textbf{Legal}} 
& \multicolumn{2}{c}{\textbf{CS}} 
& \multicolumn{2}{c}{\textbf{Mix}} \\
\cmidrule(lr){2-3}\cmidrule(lr){4-5}\cmidrule(lr){6-7}\cmidrule(lr){8-9}
& NaiveRAG & \textbf{PankRAG}
& NaiveRAG & \textbf{PankRAG}
& NaiveRAG & \textbf{PankRAG} 
& NaiveRAG & \textbf{PankRAG} \\
\midrule
% \multicolumn{9}{l}{\textbf{NaiveRAG vs PankRAG}} 
% \multicolumn{9}{2-3}{\textbf{NaiveRAG vs PankRAG}} 
% \multicolumn{9}{4-5}{\textbf{NaiveRAG vs PankRAG}} 
% \multicolumn{9}{6-7}{\textbf{NaiveRAG vs PankRAG}} \\
Comprehensiveness & 18.0\% & \textbf{82.0\%}
                   & 9.6\% & \textbf{90.4\%}
                   & 20.4\% & \textbf{79.6\%}
                   & 16.8\% & \textbf{83.2\%} \\
Diversity         & 19.6\% & \textbf{80.4\%}
                   & 7.2\% & \textbf{92.8\%}
                   & 27.6\% & \textbf{72.4\%}
                   & 8.8\% & \textbf{91.2\%} \\
logicality        & 2.8\% & \textbf{97.2\%}
                   & 8.8\% & \textbf{91.2\%}
                   & 7.6\% & \textbf{92.4\%}
                   & 17.2\% & \textbf{82.8\%} \\
relevance         & 6.0\% & \textbf{94.0\%}
                   & 11.2\% & \textbf{88.8\%}
                   & 12.8\% & \textbf{87.2\%}
                   & 19.2\% & \textbf{80.8\%} \\
coherence         & 2.8\% & \textbf{97.2\%}
                   & 8.4\% & \textbf{91.6\%}
                   & 6.8\% & \textbf{93.2\%}
                   & 16.8\% & \textbf{83.2\%} \\
Empowerment       & 12.0\% & \textbf{86.0\%}
                   & 9.6\% & \textbf{90.4\%}
                   & 16.4\% & \textbf{83.6\%}
                   & 11.6\% & \textbf{88.4\%} \\

\cmidrule(lr){2-3}\cmidrule(lr){4-5}\cmidrule(lr){6-7}\cmidrule(lr){8-9}
& GraphRAG & \textbf{PankRAG}
& GraphRAG & \textbf{PankRAG}
& GraphRAG & \textbf{PankRAG}
& GraphRAG & \textbf{PankRAG}\\
\midrule
% \multicolumn{9}{l}{\textbf{GraphRAG vs PankRAG}} \\
Comprehensiveness & 40.8\% & \textbf{59.2\%}
                   & 44.4\% & \textbf{55.6\%}
                   & 48.4\% & \textbf{53.2\%}
                   & 44.8\% & \textbf{55.2\%} \\
Diversity         & 44.8\% & \textbf{55.2\%}
                   & 39.2\% & \textbf{60.8\%}
                   & 39.6\% & \textbf{60.4\%}
                   & 43.6\% & \textbf{56.4\%} \\
logicality        & 32.8\% & \textbf{67.2\%}
                   & 43.6\% & \textbf{56.4\%}
                   & 46.4\% & \textbf{53.6\%}
                   & 30.0\% & \textbf{70.0\%} \\
relevance         & 40.4\% & \textbf{59.6\%}
                   & 38.0\% & \textbf{62.0\%}
                   & 39.6\% & \textbf{60.4\%}
                   & 44.4\% & \textbf{55.6\%} \\
coherence         & 34.4\% & \textbf{65.6\%}
                   & 38.0\% & \textbf{62.0\%}
                   & 42.4\% & \textbf{57.6\%}
                   & 31.2\% & \textbf{68.8\%} \\
Empowerment       & 39.6\% & \textbf{60.4\%}
                   & 36.4\% & \textbf{63.6\%}
                   & 44.0\% & \textbf{56.0\%}
                   & 40.0\% & \textbf{60.0\%} \\

% \midrule
% \multicolumn{9}{l}{\textbf{HyDE vs PankRAG}} \\
% Comprehensiveness & 26.0\% & \underline{74.0\%}
%                    & 41.6\% & \underline{58.4\%}
%                    & 26.8\% & \underline{73.2\%}
%                    & 40.4\% & \underline{59.6\%} \\
% Diversity         & 24.0\% & \underline{76.0\%}
%                    & 38.8\% & \underline{61.2\%}
%                    & 20.0\% & \underline{80.0\%}
%                    & 32.4\% & \underline{67.6\%} \\
% Empowerment       & 25.2\% & \underline{74.8\%}
%                    & 40.8\% & \underline{59.2\%}
%                    & 26.0\% & \underline{74.0\%}
%                    & 46.0\% & \underline{54.0\%} \\
% Overall           & 24.8\% & \underline{75.2\%}
%                    & 41.6\% & \underline{58.4\%}
%                    & 26.4\% & \underline{73.6\%}
%                    & 42.4\% & \underline{57.6\%} \\
\cmidrule(lr){2-3}\cmidrule(lr){4-5}\cmidrule(lr){6-7}\cmidrule(lr){8-9}
& LightRAG & \textbf{PankRAG}
& LightRAG & \textbf{PankRAG}
& LightRAG & \textbf{PankRAG}
& LightRAG & \textbf{PankRAG} \\
\midrule
% \multicolumn{9}{l}{\textbf{LightRAG vs PankRAG}} \\
Comprehensiveness & 44.8\% & \textbf{55.2\%}
                   & 48.4\% & \textbf{53.2\%}
                   & 35.0\% & \textbf{65.0\%}
                   & 45.2\% & \textbf{54.8\%} \\
Diversity         & 41.6\% & \textbf{58.4\%}
                   & 40.8\% & \textbf{71.6\%}
                   & 28.0\% & \textbf{72.0\%}
                   & 37.2\% & \textbf{62.8\%} \\
logicality        & 16.4\% & \textbf{83.6\%}
                   & 38.0\% & \textbf{57.2\%}
                   & 41.2\% & \textbf{58.8\%}
                   & 42.8\% & \textbf{57.6\%} \\
relevance         & 35.2\% & \textbf{64.8\%}
                   & 38.0\% & \textbf{56.0\%}
                   & 43.6\% & \textbf{56.4\%}
                   & 43.6\% & \textbf{56.4\%} \\
coherence         & 16.8\% & \textbf{83.2\%}
                   & 38.0\% & \textbf{55.6\%}
                   & 42.0\% & \textbf{58.0\%}
                   & 40.0\% & \textbf{60.0\%} \\
Empowerment       & 44.8\% & \textbf{55.2\%}
                   & 45.2\% & \textbf{58.4\%}
                   & 34.0\% & \textbf{60.0\%}
                   & 36.0\% & \textbf{64.0\%} \\
\cmidrule(lr){2-3}\cmidrule(lr){4-5}\cmidrule(lr){6-7}\cmidrule(lr){8-9}
& HippoRAG & \textbf{PankRAG}
& HippoRAG & \textbf{PankRAG}
& HippoRAG & \textbf{PankRAG}
& HippoRAG & \textbf{PankRAG} \\
\midrule
% \multicolumn{9}{l}{\textbf{LightRAG vs PankRAG}} \\
Comprehensiveness & 45.6\% & \textbf{54.4\%}
                   & 48.0\% & \textbf{52.0\%}
                   & 42.8\% & \textbf{57.2\%}
                   & 46.8\% & \textbf{53.2\%} \\
Diversity         & 42.8\% & \textbf{57.2\%}
                   & 43.2\% & \textbf{56.8\%}
                   & 36.0\% & \textbf{74.0\%}
                   & 44.0\% & \textbf{56.0\%} \\
logicality        & 31.2\% & \textbf{68.8\%}
                   & 43.6\% & \textbf{56.4\%}
                   & 42.8\% & \textbf{57.2\%}
                   & 44.4\% & \textbf{55.6\%} \\
relevance         & 40.4\% & \textbf{59.6\%}
                   & 41.2\% & \textbf{58.8\%}
                   & 47.2\% & \textbf{52.8\%}
                   & 44.8\% & \textbf{55.2\%} \\
coherence         & 32.4\% & \textbf{67.6\%}
                   & 42.4\% & \textbf{57.6\%}
                   & 39.6\% & \textbf{60.4\%}
                   & 45.2\% & \textbf{54.8\%} \\
Empowerment       & 46.0\% & \textbf{54.0\%}
                   & 46.8\% & \textbf{53.2\%}
                   & 40.8\% & \textbf{59.2\%}
                   & 41.2\% & \textbf{58.8\%} \\

\bottomrule
\end{tabular}
}
}
\vspace{-0.5cm}
\end{table}

\subsubsection{Dependency-Aware Reranking Mechanism}
% The dependency‑aware reranking mechanism is triggered when resolving a sub‑question that depends on previously resolved sub‑questions. It dynamically adjusts retrieval priorities by leveraging these resolved dependencies. Unlike conventional methods that directly incorporate resolved sub-question dependencies, risking cumulative hallucination errors across iterative reasoning steps, our mechanism mitigates such propagation by reinforcing contextually relevant content and filtering out incongruent evidence through semantic similarity validation. If no dependencies are present, the generation proceeds without reranking, preserving computational efficiency while maintaining robustness against cascading inaccuracies.

% Our dependency-aware reranking mechanism is triggered when a sub-question relies on previously resolved context. Unlike conventional methods that risk cumulative hallucinations by directly incorporating prior answers, our approach indirectly leverages these dependencies. It refines retrieval priorities by validating semantic similarity, reinforcing relevant evidence while filtering out incongruent content.
The dependency‑aware reranking mechanism is triggered when resolving a sub‑question that depends on previously resolved sub‑questions. It dynamically adjusts retrieval priorities by indirectly leveraging these resolved dependencies. Unlike conventional methods that directly incorporate resolved sub-question dependencies, risking cumulative hallucination errors across iterative reasoning steps, our mechanism mitigates such propagation by reinforcing contextually relevant content and filtering out incongruent evidence through semantic similarity validation. If no dependencies are present, the generation proceeds without reranking, preserving computational efficiency while maintaining robustness against cascading inaccuracies.

% The reranking mechanism computes a composite score by weighted summation of two components: the intrinsic quality score $ R_i $ of retrieved content and the semantic similarity $ M_i $ between retrieved content and resolved dependent sub-question answers. 
% For global search results (community summaries), the quality score is derived from precomputed metrics stored in the knowledge graph. 
% For local search results (entities and relationships), a rank-to-score transformation function converts their original ranks based on quality $ \{id_i\} $ into normalized quality scores $ \{r_i\} $. Semantic relevance is quantified using cosine similarity between the embeddings of retrieved content $ \{ori_i\} $ and resolved dependent sub-question answers$ \{sub_i\} $. 
% The intrinsic quality score $ R_i $ of retrieved content is generated and retained during the original retrieval phase~\cite{edge2024local}. The semantic similarity $ M_i $ is computed as the cosine similarity between resolved sub-question dependencies $ \{sub_i\} $ and the initially retrieved text chunks $ \{ori_i\} $. Each chunk’s similarity score is updated and then linearly combined with its original retrieval quality score to optimize prioritization.

The reranking mechanism computes a composite score by weighted summation of two components: the intrinsic quality score $ R_i $ of retrieved content and the semantic similarity $ M_i $ between retrieved content and resolved dependent sub-question answers. 
\begin{equation}
\mathcal{M}_i = \textit{Cos}(\textit{ori}_i + \textit{sub}_i)
\end{equation}
The intrinsic quality score $ R_i $ of retrieved content is generated and retained during the original retrieval phase~\cite{edge2024local}. The semantic similarity $ M_i $ is computed as the cosine similarity between resolved sub-question dependencies $ \{sub_i\} $ and the initially retrieved text chunks $ \{ori_i\} $. Each chunk’s similarity score is updated and then linearly combined with its original retrieval quality score to optimize prioritization.
The weight parameters $\alpha$ (quality) and $\beta$ (relevance) were selected by testing on the dataset(\autoref{sec:Hyperparameter}), ensuring maximal retrieval precision while mitigating hallucination propagation. This process can be formally described as follows.
% The reranking mechanism calculates a composite score for each text chunk by linearly combining its intrinsic quality $ R_i $, which is generated during the original retrieval phase~\cite{edge2024local}, with its semantic similarity $ M_i $. This similarity is computed as the cosine distance between the chunk $ \{ori_i\} $) and the answers from resolved sub-questions $ \{sub_i\} $ to optimize prioritization.
\begin{equation}
\text{Combined Score}=\alpha\times \mathcal{R}_i + \beta\times \mathcal{M}_i
\end{equation}
% The weight parameters $\alpha$ (quality) and $\beta$ (relevance) were selected by testing on the dataset(\autoref{sec:Hyperparameter}), ensuring maximal retrieval precision while mitigating hallucination propagation. 
% The weights $\alpha$ (quality) and $\beta$ (relevance) are empirically tuned (\autoref{sec:Hyperparameter}) to maximize retrieval precision while mitigating hallucination propagation.
% This process can be formally described as follows.
% \begin{equation}
% % Community detection
% \mathcal{R}_i = \textit{Transform}(\mathcal{ID}_i)
% \label{rank-to-score}
% \end{equation}

\subsubsection{Generation}
\begin{equation}
Ans = \textit{LLM}(Q,(D,K))
\end{equation}
The generation process integrates resolved dependencies $ D $ and dependency-enhanced reranking knowledge $ K $ into a consolidated context, which is utilized by the LLM to generate reliable responses.
% LLMs integrates the query $ Q $ with context($ D $,$ K $), generating responses through user‑centric adaptations and semantic coherence validation. 
% This process ensures precise alignment with the query’s underlying intent while preserving informational fidelity.

\section{Experiment}
\subsection{Experimental Setup}

 \begin{table*}
\centering
\small
\caption{Overall Performance on the SCQ Datasets}
\label{tab:specificq_res}
\resizebox{1\textwidth}{!}{  % ← 缩小到 80% 页宽
{ % 开始一个局部作用域
\renewcommand{\arraystretch}{0.2} % 将行高缩放为原来的85%
\begin{tabular}{lccccccccccccc}
\toprule

  & \multicolumn{4}{c}{Multihop-RAG} 
  & \multicolumn{4}{c}{MuSiQue} 
  & \multicolumn{4}{c}{Average} \\
\cmidrule(lr){2-5}\cmidrule(lr){6-9}\cmidrule(lr){10-13}
Method 
& \makecell{Answer \\ Relevance}  & Faithfulness & \makecell{Context \\ Recall} & \makecell{Context \\ Precision} 
& \makecell{Answer \\ Relevance}  & Faithfulness & \makecell{Context \\ Recall} & \makecell{Context \\ Precision}
& \makecell{Answer \\ Relevance}  & Faithfulness & \makecell{Context \\ Recall} & \makecell{Context \\ Precision} \\
% & \makecell{Answer \\ Relevance}  & Faithfulness & \makecell{Context \\ Recall} & \makecell{Context \\ Precision} \\
\midrule
NaiveRAG     & 66.73\% & 73.99\% & 55.17\% & 54.00\% & 66.70\% & 62.50\% & 59.41\% & 58.21\% & 66.72\% & 66.70\%  & 57.29\% & 56.11\% \\
GraphRAG     & 69.56\% & 86.80\% & 62.06\% & 61.18\% & 68.94\% & 69.73\% & 61.57\% & 60.40\% & 69.25\% & 78.27\% & 61.82\% & 60.70\% \\
LightRAG     & 72.40\% & 86.78\% & 65.18\% & 67.00\% & 67.92\% & 68.68\% & 60.99\% & 58.42\% & 68.66\% & 77.73\% & 63.08\% & 62.71\% \\
HippoRAG & 72.81\% & 89.84\% & 73.69\% & 72.35\% & 69.18\% & 75.68\% & 62.46\% & 62.19\% & 70.98\% & 82.76\% & 68.01\% & 67.27\% \\
\midrule
PankRAG    & \textbf{74.79\%} & \textbf{96.53\%} & \textbf{87.89\%} & \textbf{84.00\%} 
           & \textbf{70.62\%} & \textbf{88.75\%} & \textbf{64.99\%} & \textbf{65.35\%} 
           & \textbf{72.71\%} & \textbf{92.64\%} & \textbf{76.44\%} & \textbf{74.68\%} \\
\bottomrule
\end{tabular}
}
}
\vspace{-0.5cm}
\end{table*}

\begin{itemize}[itemsep=0pt,topsep=0pt,parsep=0pt]
     \item \textbf{Datasets} We follow the standardized preprocessing pipelines from GraphRAG\cite{edge2024local} and LightRAG\cite{guo2024lightrag} and evaluate our method on six real‑world datasets, encompassing both SCQ and ACQ. We use the Musique \cite{trivedi2022musique} and MultiHop-RAG \cite{tang2024multihop} datasets for SCQ. For ACQ, We evaluate our method on the UltraDomin dataset~\cite{DBLP:conf/naacl/QinJHZWYSLLMWB24}, selecting the four domains used (Agriculture, Legal, Computer Science, and Mix)
     \item \textbf{Baselines} We compare PankRAG with the three state-of-the-art methods, including  NaiveRAG\cite{ram-etal-2023-context}, GraphRAG\cite{edge2024local}, LightRAG\cite{guo2024lightrag} and HippoRAG\cite{jimenez2024hipporag} across all datasets.
     \item \textbf{Evaluation Metrics}  On the SCQ datasets, we employ four metrics from the Ragas framework~\cite{es-etal-2024-ragas} to assess our method and the baselines: Answer Relevance, Faithfulness, Context Recall, and Context Precision. Lacking ground-truth for the ACQ dataset, we employ GPT-4o-mini as an automated judge. We conduct blinded, pairwise comparisons between Our evaluation involves calculating win rates for our method and baselines across six dimensions. To mitigate positional bias, the presentation order of answers is alternated, and we report the averaged results. We employed GPT-4o-mini for all core tasks. For text embeddings, we utilized the BAAI/bge-m3 model.
     
     % The selection of these dimensions is informed by recent literature: Comprehensiveness, Diversity, and Empowerment are based on the evaluation framework in LightRAG~\cite{guo2024lightrag}, while Logicality, Relevance, and Coherence are sourced from the recent PathRAG~\cite{chen2025pathrag}. 
     
 \end{itemize}
\begin{figure}
    \centering
    \includegraphics[width=1\linewidth]    
    {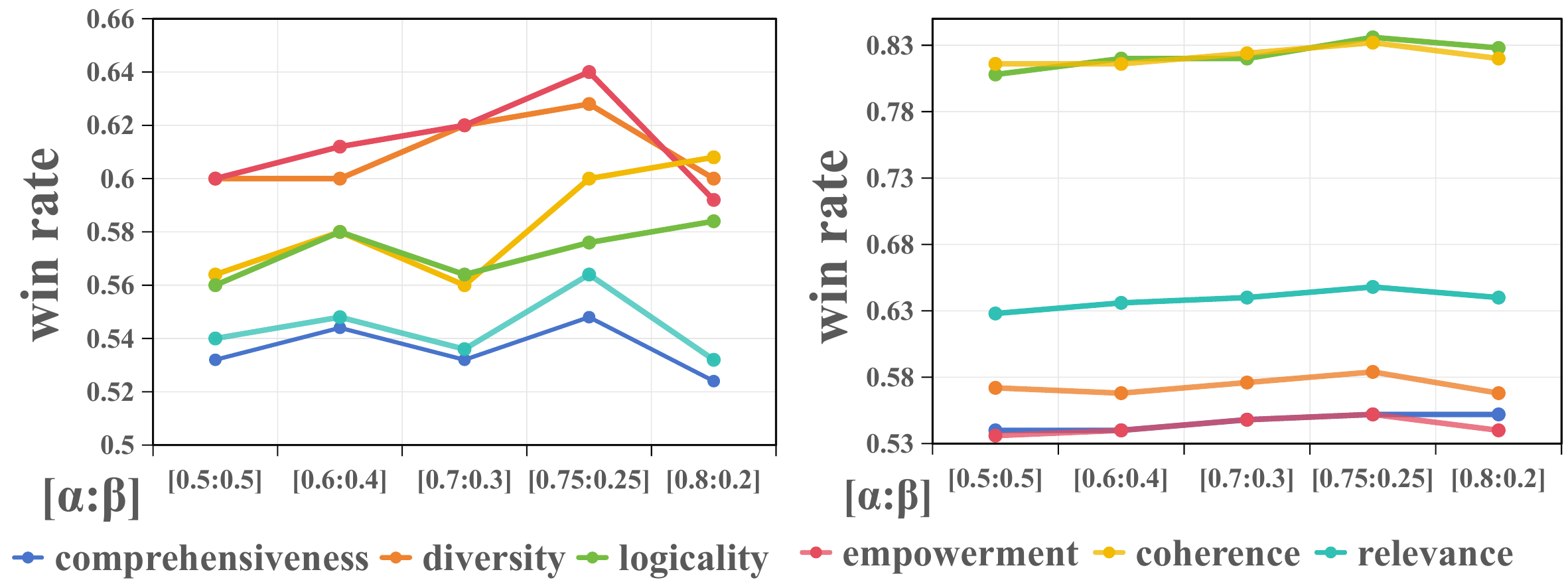}
    \vspace{-0.8cm}%%压缩图片和cap之间
    \caption{Impact of a pair of hyperparameters in PankRAG on the Mix(left) and Agriculture(right) dataset}
    \label{fig:hyperparameter_abs}
    \vspace{-0.6cm}
\end{figure}

\subsection{Overall Performance on the ACQ Datasets}
As demonstrated in \autoref{tab:abstract_res}, \textbf{PankRAG consistently surpasses baseline methods across all evaluation metrics and datasets.}

PankRAG demonstrates comprehensive superiority, achieving high average win rates across diverse datasets, ranging from 64\% to 74\%. This dominance extends across all evaluation dimensions, excelling particularly in Relevance (69.8\%) and Empowerment (69.7\%). This consistent outperformance is a direct result of its core mechanisms—globally-aware pathway resolution and dependency-aware reranking—validating its effectiveness on ACQ.

\begin{figure}
    \centering
    \includegraphics[width=1\linewidth]{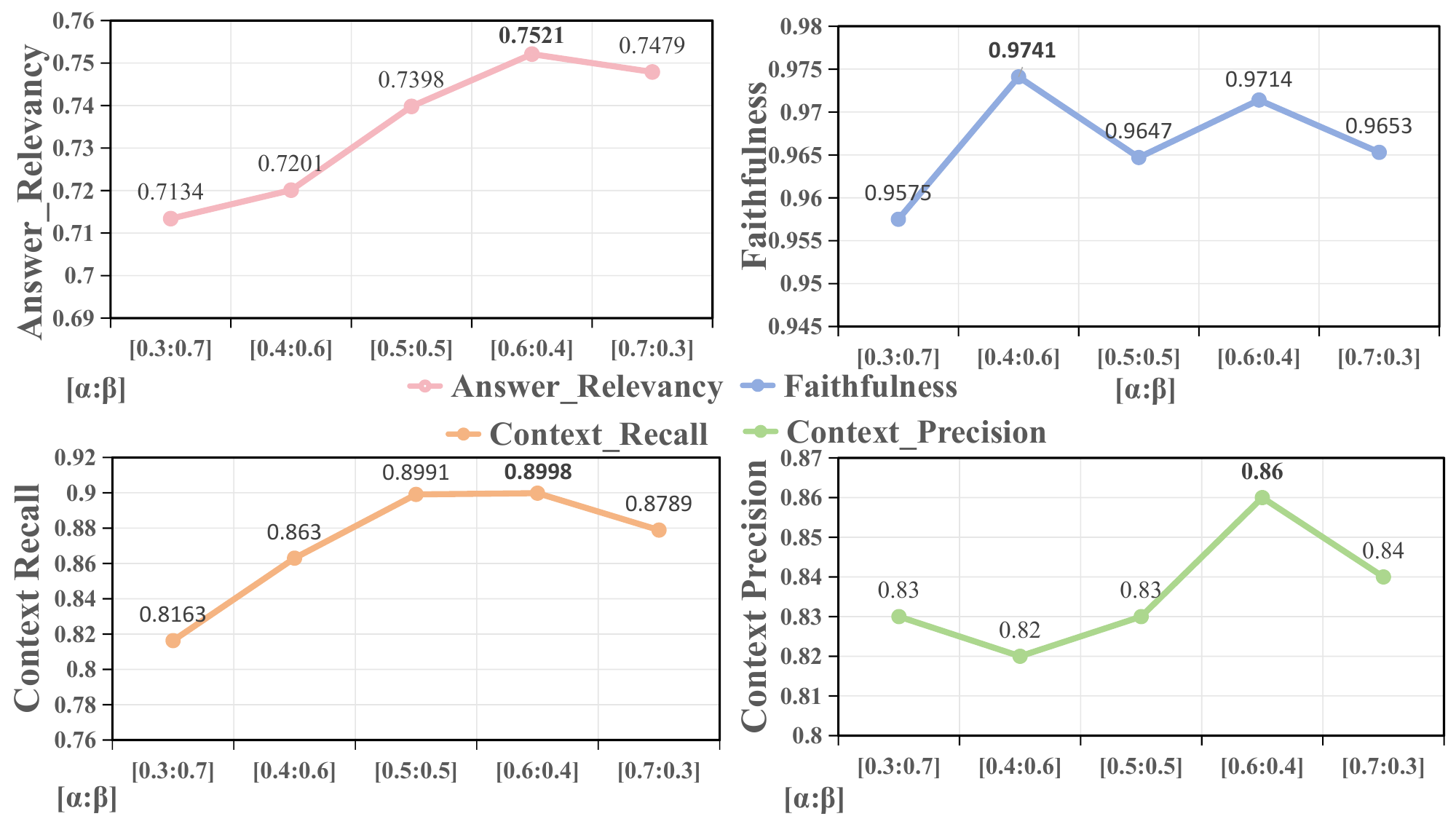}
    \vspace{-0.8cm}%%压缩图片和cap之间
    \caption{Impact of a pair of hyperparameters in PankRAG on the Multihop-RAG dataset\cite{tang2024multihop}}
    \label{fig:hyperparameter_spec}
    \vspace{-0.6cm}
\end{figure}

\begin{table*}[!t]
\centering
\small
\caption{Ablation study of PankRAG on ACQ datasets.}
\label{tab:ablations_spec}
\resizebox{1\textwidth}{!}{  % ← 缩小到 80% 页宽
{ % 开始一个局部作用域
\renewcommand{\arraystretch}{0.2} % 将行高缩放为原来的85%
\begin{tabular}{lccccccccccccc}
\toprule
  & \multicolumn{4}{c}{Multihop-RAG} 
  & \multicolumn{4}{c}{MuSiQue} 
  & \multicolumn{4}{c}{Average} \\
\cmidrule(lr){2-5}\cmidrule(lr){6-9}\cmidrule(lr){10-13}
Method 
& \makecell{Answer \\ Relevance}  & Faithfulness & \makecell{Context \\ Recall} & \makecell{Context \\ Precision} 
& \makecell{Answer \\ Relevance}  & Faithfulness & \makecell{Context \\ Precision} & \makecell{Context \\ Recall} & \makecell{Answer \\ Relevance}  & Faithfulness & \makecell{Context \\ Recall} & \makecell{Context \\ Precision} \\

\midrule

PankRAG-Pank      & 67.96\% & 88.21\% & 62.94 & 62.21 & 67.45\% & 75.16\% & 59.90\% & 58.73\% & 67.70\% & 81.69\% & 61.42\% & 60.47\% \\
PankRAG-Ank     & 70.21\% & 94.45 & 84.31 & 67.65 &  69.02\% & 81.13\% & 62.27\% & 61.23\% & 69.61\% & 87.79\% & 73.29\% & 68.63\% \\
% PankRAG & 0.7479 & 0.9653 & 0.8789 & 0.8400 & 0.7062 & 0.8875 & 0.6499 & 0.6535 & 46.70 & 59.98 & 52.17 & 60.79 \\
\midrule
PankRAG    & \textbf{74.79\%} & \textbf{96.53\%} & \textbf{87.89\%} & \textbf{84.00\%} 
           & \textbf{70.62\%} & \textbf{88.75\%} & \textbf{64.99\%} & \textbf{65.35\%} 
           & \textbf{72.71\%} & \textbf{92.64\%} & \textbf{76.44\%} & \textbf{74.68\%} \\
\bottomrule
\end{tabular}
}
}
\vspace{-0.5cm}
\end{table*}
 \subsection{Overall Performance on the SCQ Datasets}
As demonstrated in \autoref{tab:specificq_res},
\textbf{PankRAG consistently achieves the best performance across all experimental settings.} PankRAG establishes its superiority by outperforming baselines by 11-17\% on average. This advantage is driven by substantial gains across key metrics, including Faithfulness (10-26\%), Context Precision(7-18\%), Context Recall (8-19\%), and Answer Relevance(2-6\%), underscoring its enhanced efficacy in resolving SCQs.
 
% Compared to NaiveRAG, GraphRAG, and LightRAG, PankRAG achieves average metric improvements of 17.42\%, 11.61\%, 11.08\%. Averaged over multiple datasets, the results are as follows. PankRAG achieves approximately 5.99\% higher than NaiveRAG, 3.46\% higher than GraphRAG, and 4.05\% higher than LightRAG in Answer Relevance; 25.94\% higher than NaiveRAG, 14.37\% higher than GraphRAG, and 14.91\% higher than LightRAG in Faithfulness; 19.15\% higher than NaiveRAG, 14.62\% higher than GraphRAG, and 13.36\% higher than LightRAG in Context Recall; and 18.57\% higher than NaiveRAG, 13.98\% higher than GraphRAG, and 11.97\% higher than LightRAG in Context Precision.
% These results demonstrate PankRAG’s superiority over prior graph-based RAG methods in handling concrete complex problems.

% PankRAG demonstrates a consistent advantage over all baselines with average performance gains of 17.42\%, 11.61\%, and 11.08\%, respectively.
% This superiority is reflected across all key metrics. Notably, PankRAG achieves substantial gains in Faithfulness (14\%-26\% improvement over baselines) and enhances Context Precision and Context Recall by 12-19\%. Answer Relevance also sees a marked improvement. These results validate PankRAG's enhanced capability for resolving SCQ.

Notably, GraphRAG\cite{edge2024local} and LightRAG\cite{guo2024lightrag} show minimal improvements in Context Recall and Context Precision  (4\%-6\% over NaiveRAG), indicating negligible enhancement in retrieved context quality. This aligns with the identified challenge that LLMs struggle to interpret hidden relationships or information in SCQ, leading to misaligned retrieval targets. 
% While these methods enhance retrieval via graph structures, their limited context quality gains highlight unresolved limitations.
In contrast, PankRAG’s significant improvements in Context Recall and Context Precision suggest effective mitigation of this issue, attributed to its globally-aware pathway resolution generation module and dependency-aware reranking mechanism. It enables LLMs to better capture hidden relationships and thereby improve context relevance. Concurrent enhancements in Faithfulness and Answer Relevance further indicate reduced hallucination risks and higher answer quality, underscoring PankRAG’s comprehensive superiority over existing graph-based RAG methods.
\begin{table}[!t]
\centering
\caption{Ablation study of PankRAG on ACQ datasets.}
\label{tab:ablation_abs}
\resizebox{1\columnwidth}{!}{  % ← 缩小到 80% 页宽
{ % 开始一个局部作用域
\renewcommand{\arraystretch}{0.3} % 将行高缩放为原来的85%
\setlength{\tabcolsep}{0.5pt}
\begin{tabular}{lcccccccc}
\toprule
& \multicolumn{2}{c}{\textbf{Agriculture}} 
& \multicolumn{2}{c}{\textbf{Legal}} 
& \multicolumn{2}{c}{\textbf{CS}}
& \multicolumn{2}{c}{\textbf{Mix}} \\
\cmidrule(lr){2-3}\cmidrule(lr){4-5}\cmidrule(lr){6-7}\cmidrule(lr){8-9}
& w/o Pank & \textbf{w/o Ank}
& w/o Pank & \textbf{w/o Ank}
& w/o Pank & \textbf{w/o Ank} 
& w/o Pank & \textbf{w/o Ank} \\
\midrule
% \multicolumn{9}{l}{\textbf{NaiveRAG vs PankRAG}} 
% \multicolumn{9}{2-3}{\textbf{NaiveRAG vs PankRAG}} 
% \multicolumn{9}{4-5}{\textbf{NaiveRAG vs PankRAG}} 
% \multicolumn{9}{6-7}{\textbf{NaiveRAG vs PankRAG}} \\
Comprehensiveness & 46.8\% & \textbf{53.2\%}
                   & 34.8\% & \textbf{55.2\%}
                   & 47.2\% & \textbf{52.8\%}
                   & 46.0\% & \textbf{54.0\%} \\
Diversity         & 42.3\% & \textbf{58.8\%}
                   & 38.0\% & \textbf{62.0\%}
                   & 36.2\% & \textbf{54.8\%}
                   & 33.6\% & \textbf{56.4\%} \\
logicality        & 44.4\% & \textbf{55.6\%}
                   & 44.0\% & \textbf{56.0\%}
                   & 44.8\% & \textbf{55.2\%}
                   & 45.2\% & \textbf{54.8\%} \\
relevance         & 45.2\% & \textbf{54.8\%}
                   & 44.8\% & \textbf{55.2\%}
                   & 46.4\% & \textbf{53.6\%}
                   & 46.0\% & \textbf{54.0\%} \\
coherence         & 46.8\% & \textbf{53.2\%}
                   & 43.8\% & \textbf{7.2\%}
                   & 46.4\% & \textbf{53.6\%}
                   & 45.2\% & \textbf{54.8\%} \\
Empowerment       & 56.0\% & \textbf{54.0\%}
                   & 41.2\% & \textbf{58.8\%}
                   & 46.0\% & \textbf{54.0\%}
                   & 43.6\% & \textbf{54.4\%} \\

\cmidrule(lr){2-3}\cmidrule(lr){4-5}\cmidrule(lr){6-7}\cmidrule(lr){8-9}
& w/o Ank & \textbf{PankRAG}
& w/o Ank & \textbf{PankRAG}
& w/o Ank & \textbf{PankRAG}
& w/o Ank & \textbf{PankRAG} \\
\midrule
% \multicolumn{9}{l}{\textbf{GraphRAG vs PankRAG}} \\
Comprehensiveness & 41.6\% & \textbf{58.4\%}
                   & 46.0\% & \textbf{54.0\%}
                   & 46.0\% & \textbf{54.0\%}
                   & 39.2\% & \textbf{55.2\%} \\
Diversity         & 46.4\% & \textbf{53.6\%}
                   & 44.8\% & \textbf{55.2\%}
                   & 41.2\% & \textbf{58.8\%}
                   & 30.8\% & \textbf{56.4\%} \\
logicality        & 33.2\% & \textbf{66.8\%}
                   & 43.2\% & \textbf{56.8\%}
                   & 44.8\% & \textbf{55.2\%}
                   & 32.4\% & \textbf{68.8\%} \\
relevance         & 41.6\% & \textbf{58.4\%}
                   & 46.0\% & \textbf{54.0\%}
                   & 45.6\% & \textbf{54.4\%}
                   & 32.4\% & \textbf{56.4\%} \\
coherence         & 36.8\% & \textbf{63.2\%}
                   & 43.0\% & \textbf{57.0\%}
                   & 43.6\% & \textbf{56.4\%}
                   & 32.4\% & \textbf{60.0\%} \\
Empowerment       & 37.2\% & \textbf{62.8\%}
                   & 43.4\% & \textbf{55.6\%}
                   & 44.0\% & \textbf{56.0\%}
                   & 42.4\% & \textbf{58.8\%} \\

% \midrule
% \multicolumn{9}{l}{\textbf{HyDE vs PankRAG}} \\
% Comprehensiveness & 26.0\% & \underline{74.0\%}
%                    & 41.6\% & \underline{58.4\%}
%                    & 26.8\% & \underline{73.2\%}
%                    & 40.4\% & \underline{59.6\%} \\
% Diversity         & 24.0\% & \underline{76.0\%}
%                    & 38.8\% & \underline{61.2\%}
%                    & 20.0\% & \underline{80.0\%}
%                    & 32.4\% & \underline{67.6\%} \\
% Empowerment       & 25.2\% & \underline{74.8\%}
%                    & 40.8\% & \underline{59.2\%}
%                    & 26.0\% & \underline{74.0\%}
%                    & 46.0\% & \underline{54.0\%} \\
% Overall           & 24.8\% & \underline{75.2\%}
%                    & 41.6\% & \underline{58.4\%}
%                    & 26.4\% & \underline{73.6\%}
%                    & 42.4\% & \underline{57.6\%} \\

\bottomrule
\end{tabular}
}
}
\vspace{-0.6cm}
\end{table}

\subsection{Hyperparameter Analysis}
\label{sec:Hyperparameter}
% The parameters $\alpha$ and $\beta$ are designed to optimize the original quality score by integrating a novel similarity metric, constrained by $\alpha$ + \(\beta\) = 1 to ensure the normalized score range.

The parameters $\alpha$ and $\beta$ (constrained by $\alpha$ + $\beta$ = 1 to ensure the normalized score range) balance the original quality score against a novel similarity metric derived from resolved dependencies. As shown in \autoref{fig:hyperparameter_abs} and \autoref{fig:hyperparameter_spec}, Experimental results demonstrate that \textbf{the optimal parameters for SCQ are $\alpha$ \textgreater 0.6 $\beta$ \textless 0.4, whereas for ACQ, $\alpha$ = 0.75 and $\beta$ = 0.25 , with both configurations maximizing retrieval performance.}

For SCQ, the optimal performance is achieved with $\alpha$ = 0.6 and $\beta$ = 0.4. This configuration represents a critical trade-off. A higher weight on $\alpha$ ($\alpha$ \textgreater 0.6) results in insufficient reinforcement from sub-question answers, while a lower weight ($\alpha$ \textless 0.6) leads to an over-emphasis that introduces negative hallucinations, degrading performance. The 0.6:0.4 ratio, therefore, strikes the ideal balance between leveraging sub-questions for context validation and mitigating the risk of performance degradation, yielding the best overall retrieval results. For ACQ, our experiments identified a different optimal configuration of $\alpha$ = 0.75 and $\beta$ = 0.25. This setting was determined as it consistently maximized retrieval performance and achieved the highest win rates against baseline models.

\subsection{Ablation Studies}
This section's ablation studies validate the hierarchical contributions of our two core components: globally-aware pathway resolution (Path) and dependency-aware reranking (Ank).

As shown in \autoref{tab:ablations_spec} and \autoref{tab:ablation_abs}, the results demonstrate that \textbf{the Path module is a foundational component crucial for enhancing retrieval quality}. The model variant equipped with Path (PankRAG-Ank) achieves significant improvements in Context Precision and Context Recall over the variant without it (PankRAG-Pank). This finding underscores the necessity of the globally-aware pathway resolution for establishing a high-quality context. Furthermore, \textbf{the Ank mechanism provides a critical, synergistic enhancement on top of the Path framework}. The full PankRAG system, which incorporates the Ank module, shows consistent superiority over the PankRAG-Ank variant across all evaluated metrics. This highlights the critical role of the dependency-aware reranking mechanism in optimizing and refining the final results.

\section{Conclusion}
In this paper, we propose PankRAG, a novel RAG framework designed to enhance LLMs’ capacity to accurately interpret and retrieve information for complex queries. 
PankRAG combines a globally aware pathway resolution module with a dependency‐aware reranking mechanism.
By executing a globally coherent reasoning pathway and augmenting retrieved
content through dependency‐aware reranking, our framework captures and resolves hidden relationships and information in both SCQ and ACQ, significantly enhancing retrieval performance and response quality.

\vfill\pagebreak

% \section{REFERENCES}
% \label{sec:refs}

% List and number all bibliographical references at the end of the
% paper. The references can be numbered in alphabetic order or in
% order of appearance in the document. When referring to them in
% the text, type the corresponding reference number in square
% brackets as shown at the end of this sentence \cite{C2}. An
% additional final page (the fifth page, in most cases) is
% allowed, but must contain only references to the prior
% literature.

% Please follow the IEEE Citation Guidelines, \url{https://ieee-dataport.org/sites/default/files/analysis/27/IEEE\%20Citation\%20Guidelines.pdf} for formatting of references.

% References should be produced using the bibtex program from suitable
% BiBTeX files (here: strings, refs, manuals). The IEEEbib.bst bibliography
% style file from IEEE produces unsorted bibliography list.
% -------------------------------------------------------------------------
\bibliographystyle{IEEEbib}
\bibliography{main}

\end{document}